\documentclass[sigconf]{acmart}

\makeatletter                   %1
\def\mdseries@tt{m}             %1
\makeatother                    %1

\usepackage[draft=true]{minted} %2

\usepackage{amsfonts}
\usepackage{amssymb,amsbsy}
\usepackage{amsmath}
\usepackage{amsfonts, amscd, amsthm}
\usepackage{algorithm, algorithmic}
\usepackage{color}
\usepackage{url}
\usepackage{verbatim}
\usepackage{setspace}
\usepackage[abs]{overpic}
\usepackage{graphicx}
\usepackage{subfigure}
\usepackage{bbm, bm}
\usepackage{accents}
\usepackage{fullpage}
\usepackage{dsfont}

\newtheorem*{example*}{Example}
\numberwithin{equation}{section}

\newtheorem{example}{Example}

\renewcommand{\bm}{\boldsymbol}
\newcommand{\mc}{\mathcal}

\newcommand{\norm}[2]{\left\| #1 \right\|_{#2}}
\newcommand{\reals}{\mathbb{R}}

\newcommand{\set}[1]{\left\{ #1 \right\}}

\newlength{\dhatheight}

\usepackage{booktabs} % For formal tables
\usepackage{multirow}
\usepackage{array}
\usepackage{multicol}
\usepackage{arydshln}

\newcolumntype{L}[1]{>{\raggedright\let\newline\\\arraybackslash\hspace{0pt}}m{#1}}
\newcolumntype{C}[1]{>{\centering\let\newline\\\arraybackslash\hspace{0pt}}m{#1}}
\newcolumntype{R}[1]{>{\raggedleft\let\newline\\\arraybackslash\hspace{0pt}}m{#1}}

\DeclareMathOperator*{\rargmax}{r-arg\,max}
\DeclareMathOperator*{\sargmin}{s-arg\,min}

\setcopyright{rightsretained}
\acmDOI{}

\acmISBN{}

\setcopyright{rightsretained}
\acmConference[SIGIR 2018 eCom]{ACM SIGIR Workshop on eCommerce}{July 2018}{Ann Arbor, Michigan, USA}
\acmYear{2018}
\copyrightyear{2018}

\begin{document}
\title{Towards Practical Visual Search Engine \\Within Elasticsearch}
%\titlenote{Produces the permission block, and
 % copyright information}

\author{Cun (Matthew) Mu}
%\authornote{This author is the
%	one who did all the really hard work.}
\affiliation{%
  \institution{Jet.com/Walmart Labs}
  \city{Hoboken}
  \state{NJ}
}
\email{matthew.mu@jet.com}

\author{Jun (Raymond) Zhao}
\affiliation{%
	\institution{Jet.com/Walmart Labs}
	\city{Hoboken}
	\state{NJ}
}
\email{raymond@jet.com}

\author{Guang Yang}
\affiliation{%
	\institution{Jet.com/Walmart Labs}
	\city{Hoboken}
	\state{NJ}
}
\email{guang@jet.com}

\author{Jing  Zhang}
\affiliation{%
	\institution{Jet.com/Walmart Labs}
	\city{Hoboken}
	\state{NJ}
}
\email{jing@jet.com}

\author{Zheng (John) Yan}
\affiliation{%
	\institution{Jet.com/Walmart Labs}
	\city{Hoboken}
	\state{NJ}
}
\email{john@jet.com}


\begin{abstract}
In this paper, we describe our end-to-end content-based image retrieval system built upon Elasticsearch, a well-known and popular textual search engine. As far as we know, this is the first time such a system has been implemented in eCommerce, and our efforts have turned out to be highly worthwhile. We end up with a novel and exciting visual search solution that is extremely easy to be deployed, distributed, scaled and monitored in a cost-friendly manner. Moreover, our platform is intrinsically flexible in supporting multimodal searches, where visual and textual information can be jointly leveraged in retrieval. 

The core idea is to encode image feature vectors into a collection of string tokens in a way such that closer vectors will share more string tokens in common. By doing that, we can utilize Elasticsearch to efficiently retrieve similar images based on similarities within encoded sting tokens. As part of the development, we propose a novel vector to string encoding method, which is shown to substantially outperform the previous ones in terms of both precision and latency. 

First-hand experiences in implementing this Elasticsearch-based platform are  extensively addressed, which should be valuable to practitioners also interested in building visual search engine on top of Elasticsearch.
\end{abstract}


%https://dl.acm.org/ccs/ccs_flat.cfm
\begin{CCSXML}
	<ccs2012>
	<concept>
	<concept_id>10002951.10003317.10003371.10003386.10003387</concept_id>
	<concept_desc>Information systems~Image search</concept_desc>
	<concept_significance>500</concept_significance>
	</concept>
	<concept>
	<concept_id>10010405.10003550.10003555</concept_id>
	<concept_desc>Applied computing~Online shopping</concept_desc>
	<concept_significance>500</concept_significance>
	</concept>
	</ccs2012>
\end{CCSXML}

\ccsdesc[500]{Information systems~Image search}
\ccsdesc[500]{Applied computing~Online shopping}
\keywords{Elasticsearch, visual search, content-based image retrieval, multimodal search, eCommerce}

\maketitle

\section{Introduction}

\begin{figure*}[h]
	\fbox{\includegraphics[width=0.98\textwidth]{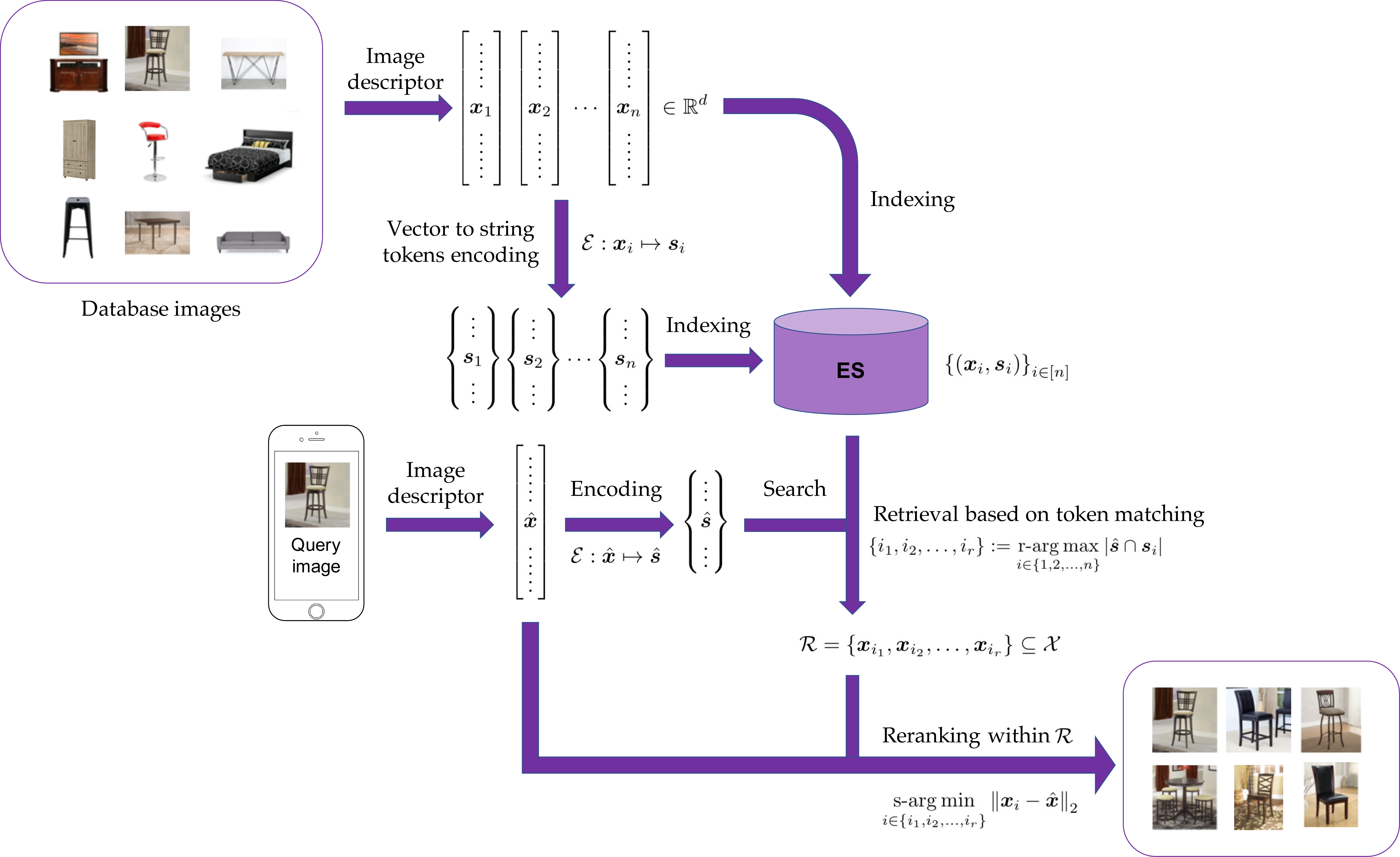}}
	\caption{{\bf Pipeline of our visual search system within Elasticsearch.} \normalfont{The image vectors and their encoded string tokens are indexed together into Elasticsearch. At search time, the query vector $\hat{\bm x}$ will be first encoded into string tokens $\hat{\bm s}$, based on which a small candidate set $\mc R$ is retrieved. We will then re-rank vectors in $\mc R$ according to their exact Euclidean distances with $\hat{\bm x}$}, and output the top ones as our final visual search outcome.} \label{fig: pipeline}
\end{figure*}

Elasticsearch \cite{gormley2015elasticsearch}, built on top of Apache Lucene library \cite{mccandless2010lucene, bialecki2012apache, lucenelib2018}, is an {\em open-source}, {\em real-time}, {\em distributed} and {\em multi-tenant} textual search engine. Since its first release in February 2010, Elasticsearch has been widely adopted by eCommerce websites (e.g., Ebay, Etsy, Jet, Netflix, Grubhub) to successfully help customers discover products based on the textual queries they requested \cite{esusers2018, es2018} .

But {\em a picture is more than often worth a thousand words}. With the explosive usage of phone cameras, {\em content-based  image retrieval} \cite{datta2008image} is increasingly demanded from customers. Especially for categories like furniture, fashion and lifestyle (where buying decisions are largely influenced  by products' visual appealingness), uploading a picture of the product they like could be substantially more specific, expressive and straightforward than elaborating it into abstract textual description. 

Finding images relevant with the uploaded picture tends to be much more involved and vaguer than retrieving documents matching keywords \cite{salton1975vector, tfidf2018, robertson2009probabilistic} typed into the search box, as words (by themselves) are substantially more semantic and meaningful than image pixel values. 
%The task of content-based  image retrieval could be substantially more involved and vaguer than text retrieval (using vector space model \cite{salton1975vector}). 
Fortunately, modern AI techniques, especially the ones developed in the field of deep learning \cite{bengio2009learning, goodfellow2016deep}, have made incredible strides in image feature extraction \cite{raina2007self, mesnil2011unsupervised, le2012building, donahue2014decaf, oquab2014learning, zeiler2014visualizing, razavian2014cnn, yosinski2014transferable} to embed images as points in {\em high-dimensional} Euclidean space, where similar images are located nearby. So, given a query image, we can simply retrieve its visually similar images by {\em finding its nearest neighbors in this high-dimensional feature space}. However,  Elasticsearch, as an {\em inverted-index-based search engine}, is not much empowered to accomplish this mathematically straightforward operation in an efficient manner (though efforts \cite{gennaro2010approach, esmdp2016, lux2016lire, esndr2017, nk2018} have been made successfully in finding nearest neighbors over spaces of much lower dimension), which significantly limits the applicability of its nicely designed engineering system as well as the huge volume of product metadata already indexed into its database (for textual search). The gist of the paper is to conquer this difficulty, and thus make it feasible to conduct visual search within Elasticsearch.

In this paper, we describe our end-to-end visual search platform built upon Elasticsearch. As far as we know, this is {\em the first attempt} to achieve this goal and our efforts turn out to be quite worthwhile. By taking advantage of the mature engineering design from Elasticsearch, we end up with a visual search solution that is extremely easy to be {\em deployed}, {\em distributed}, {\em scaled} and {\em monitored}. Moreover, due to Elasticsearch's disk-based (and partially memory cached) inverted index mechanism, our system  is quite {\em cost-effective}. In contrast to many existing systems (using {\em hashing-based} \cite{andoni2006near, torralba2008small, weiss2009spectral, wang2010semi, gong2013iterative, he2013k, liu2014discrete} or {\em quantization-based} \cite{jegou2011product, jegou2011searching, jegou2012aggregating, ge2013optimized,kalantidis2014locally} {\em approximate nearest neighbor} (ANN) methods), we do not need to load those millions of (high-dimensional and dense) image feature vectors into RAM, one of the most expensive resources in large-scale computations. Furthermore, by integrating textual search and visual search into one engine, both types of product information can now be shared and utilized seamlessly in a single index. This paves a coherent way to support {\em multimodal searches}, allowing customers to express their interests in a variety of textual requests (e.g., keywords, brands, attributes, price ranges) jointly with visual queries, at which most of existing visual search systems fall short (if not impossible).

Since the image preprocessing step and the image feature extraction step involved in our system are standard and  independent of Elasticsearch, in this paper we address more towards how we empower Elasticsearch to retrieve close image feature vectors, i.e., {\em the Elasticsearch-related part of the visual system}. Our nearest neighbor retrieval approach falls under the general framework recently proposed by \citet{rygl2017semantic}. The core idea is to create {\em text documents} from {\em image feature vectors} by encoding each vector into a collection of string tokens in a way such that {\em closer vectors will share more string tokens in common}. This enables Elasticsearch to  approximately retrieve neighbors in image feature space based on their encoded textual  similarities. The quality of the encoding procedure (as expected) is extremely critical to the success of this approach. In the paper, we propose a noval  scheme called {\em subvector-wise clustering  encoder}, which substantially outperforms the element-wise rounding one proposed and examined by \citet{rygl2017semantic} and  \citet{ruuvzivckaflexible}, in terms of both {\em precision} and {\em latency}. Note that our methodology should be generally applicable to any full-text search engine (e.g., Solr \cite{smiley2015apache}, Sphinx \cite{aksyonoff2011introduction}) besides Elasticsearch, but in the paper we do share a number of Elasticsearch-specific implementation tips based on our first-hand experience, which should be valuable to practitioners interested in building their own visual search system on top of Elasticsearch.

The rest of the paper is organized as follows. In Section 2, we describe the general pipeline of our visual search system, and highlight a number of engineering tweaks we found useful when implementing the system on Elasticsearch. In Section 3 and 4, we focus on how to encode an image feature vector into a collection of string tokens---the most crucial part in setting up the system. In Section 3, we first review the element-wise rounding encoder and address its drawbacks. As a remedy, we propose a new encoding scheme called subvector-wise clustering encoder, which is empirically shown in Section 4 to much outperform the element-wise rounding one.

\section{General Framework of Visual Search within Elasticsearch}
The whole pipeline of our visual search engine is depicted in Figure \ref{fig: pipeline}, which primarily consists of two phases: indexing and searching.

%We assume each image is represented as a (potentially high-dimensional and dense) numerical vector, which encapsulates local and global features in image, through an image descriptor (e.g., SIFT descriptor, CNN descriptor). Then the task of visual search is reduced to conducting nearest neighbors search. To achieve that within Elasticsearch, we adopt the framework recently proposed by \citet{rygl2017semantic}. The core idea is to create {\em text documents} from numerical vectors by encoding each vector into a collection of string tokens in a way such that closer vectors will share more string tokens in common, so that we can leverage the fulltext search engine to retrieve similar ones. The whole pipeline is illustrated in Figure \ref{fig: pipeline}, which primarily consists of two phases: indexing and search.

\paragraph{Indexing.} Given image feature vectors 
\begin{flalign}
\mc X:= \set{\bm x_1, \bm x_2, \ldots, \bm x_n} \subseteq \reals^d,
\end{flalign}
we will first encode them into string tokens 
\begin{flalign}
\mc S := \set{\bm s_1, \bm s_2, \ldots, \bm s_n},
\end{flalign}
where $\bm s_i := \mc E(\bm x_i)$ for some encoder $\mc E(\cdot)$ converting a $d$-dimensional vector into a collection of string tokens of cardinality $m$. The original numerical vectors $\mc X$ and encoded tokens $\mc S$, together with their textual  metadata (e.g, product titles, prices, attributes), will be all indexed into the Elasticsearch database, to wait for being searched.

\paragraph{Searching.} Conceptually, the search phase consists of two steps: {\em retrieval} and {\em reranking}. Given a query vector $\hat{\bm x}$, we will first encode it into $\hat{\bm s}:= \mc E(\hat{\bm x})$ via the same encoder used in indexing, and retrieve $r$ ($r \ll n$) most similar vectors $\mc R := \set{\bm x_{i_1}, \bm x_{i_2}, \ldots, \bm x_{i_r}}$ as candidates based on the overlap between the string token set $\hat{\bm s}$ and the ones in $\set{\bm s_1, \bm s_2, \ldots, \bm s_n}$, i.e.,
\begin{flalign}\label{eqn: retrieval}
\set{i_1, i_2, \ldots, i_r} = \rargmax_{i \in \set{1, 2, \ldots, n}} \vert \hat{\bm s} \cap \bm s_i \vert.
\end{flalign} 
We will then re-rank vectors in the candidate set $\mc R$ according to their {\em exact} Euclidean distances with respect to the query vector $\hat{\bm x}$, and choose the top-$s$ ($s\le r$) ones as the final visual search result to output, i.e.,
\begin{flalign}\label{eqn: rerank}
\sargmin_{i \in \set{i_1, i_2, \ldots, i_r}}  \norm{\bm x_i - \hat{\bm x}}{2}.
\end{flalign} 

As expected, the choice of $\mc E(\cdot)$ is extremely critical to the success of the above approach. A good encoder $\mc E(\cdot)$  should encourage image feature vectors closer in Euclidean distance to share more string tokens in common, so that the retrieval set $\mc R$ obtained from the optimization problem \eqref{eqn: retrieval} could contain enough meaning candidates to be fed into the exact search in \eqref{eqn: rerank}. We will elaborate and compare different choices of encoders in the next two sections (Section \ref{sec: encoder}\&\ref{sec: experiment}).

\paragraph{Implementation.} In this part, we will address how we implement the retrieval and reranking steps in the searching phase efficiently within just one JSON-encoded request body (i.e., JSON \ref{json: search_request_body}), which instructs the Elasticsearch server to compute \eqref{eqn: retrieval} and \eqref{eqn: rerank} and then return the visual search result in a desired order (via Elasticsearch's RESTful API over HTTP). 

For the retrieval piece, we construct a {\em function score query} \cite{esfsq2018} to rank database images based on  \eqref{eqn: retrieval}. Specifically, our function score query (lines 3-29 in JSON \ref{json: search_request_body}) consists of $m$ score functions, each of which is a {\em term filter} \cite{estq2018} (e.g., lines 6-14 in JSON \ref{json: search_request_body}) to check whether the encoded feature token $\hat{ s}_i$ from the query image is being matched or not. With all the $m$ scores being summed up (line 26 in JSON \ref{json: search_request_body}) using the same weight (e.g., lines 13 and 23 in JSON \ref{json: search_request_body}), the ranking score for the database images are calculated exactly as the number of feature tokens they overlap with the ones in $\hat{\bm s}$.

%For retrieval, we use the \textit{Function Score Query} API of Elasticseach, where each string token in the encoded query vector becomes a score function with a \textit{Term} filter\footnote{https://www.elastic.co/guide/en/elasticsearch/reference/6.1/query-dsl-term-query.html}. All functions  share the same \verb|weight| parameter and their scores are summed so that documents are ranked according to the number of filtering queries matched.

For the reranking piece, our initial trial is to fetch the top-$r$ image vectors from the retrieval step, and calculate \eqref{eqn: rerank} to re-rank them outside Elasticsearch. But this approach prevents our visual system from being an end-to-end one within Elasticsearch, and thus makes it hard to leverage many useful microservices (e.g., pagination) provided by Elasticsearch. More severely, this vanilla approach introduces substantial latency in communication as thousands of high-dimensional and dense image embedding vectors have to be transported out of Elasticsearch database. As a remedy, we design a {\em query rescorer} \cite{esrs2018} (lines 30-52 in JSON \ref{json: search_request_body}) within Elasticsearch to execute a second query on the top-$r$ database image vectors returned from the function score query, to tweak their scores and re-rank them based on their exact Euclidean distances with the query image vector. In specific, we implement a custom {\em Elasticsearch plugin} \cite{esplugin2018}  (lines 35-47 in JSON \ref{json: search_request_body}) to compute the {\em negation} of the Euclidean distance between query image vector and the one from  database. As Elasticsearch will rank the result based on the ranking score from high to low, the output will be in the desired order from the smallest distance to the largest one.

%For reranking, we leverage the \textit{Rescore} API\footnote{https://www.elastic.co/guide/en/elasticsearch/reference/6.1/search-request-rescore.html}. We set the \verb|window_size| parameter to \(r\) to rerank only the top \(r\) retrieved documents. We implemented a custom Elasticsearch plugin\footnote{https://www.elastic.co/guide/en/elasticsearch/reference/6.1/modules-plugins.html} to compute the Euclidean distance between query and document vectors. Doing so avoids returning hundreds documents from Elasticsearch and rerank on the client side, and allows us to easily handle pagination.

\renewcommand{\listingscaption}{JSON}
\begin{listing}[h!]
\begin{minted}[linenos=true]{json}
{
  "size": s,
  "query": {
    "function_score": {
      "functions": [
      {
        "filter": {
          "term": {
            "image_encoded_tokens": 
              "query_encoded_token_1"
          }
        },
        "weight": 1 
      },
      ...,
      {
        "filter": {
          "term": {
            "image_encoded_tokens": 
              "query_encoded_token_m"
            }
          },
          "weight": 1
        }
      ],
      "score_mode": "sum", 
      "boost_mode": "replace"
    }
  },
  "rescore": {
    "window_size": r,
    "query": {
      "rescore_query": {
        "function_score": {
        "script_score": {
          "script": {
            "lang": "custom_scripts",
            "source": "negative_euclidean_distance",
            "params": {
              "vector_field": "image_actual_vector",
              "query_vector": 
                [0.1234, -0.2394, 0.0657, ...]
            }
          }
        },
        "boost_mode": "replace"
      }
    },
    "query_weight": 0,
    "rescore_query_weight": 1
    }
  }
}
\end{minted}
\caption{Request body for visual search in Elasticsearch 6.1}
\label{json: search_request_body}
\end{listing}

\begin{figure}[ht]
	\centerline{\includegraphics[width=3.1in]{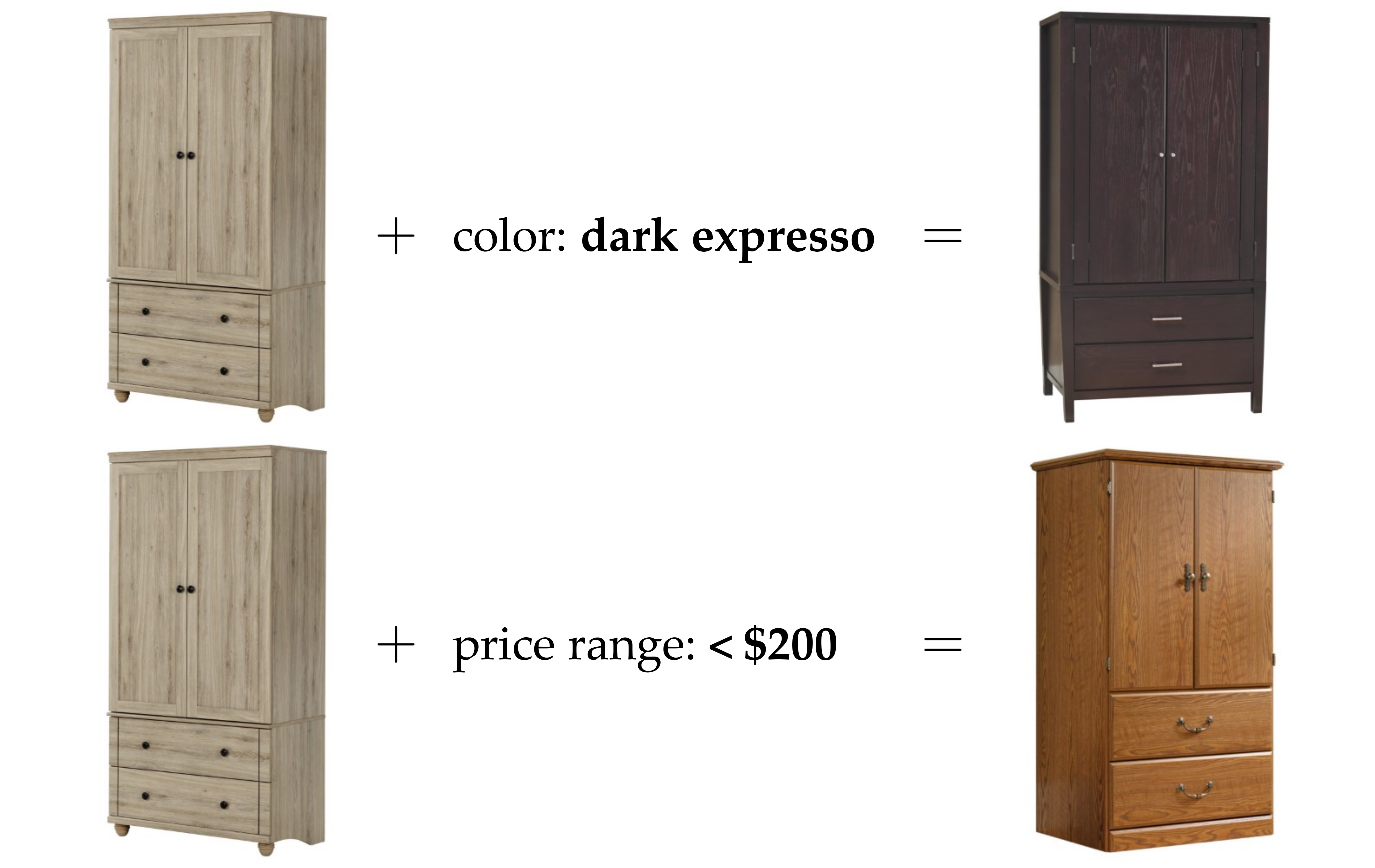}}
	\caption{{\bf Illustration of multimodal search.} \normalfont{Armoire is searched using image query jointly with color/price range specified by the customer. Our Elasticsearch-based visual search engine can be easily tailored to handle complicated business requests like the above by adding filters (e.g., {\em term filter} \cite{estq2018}, {\em range filter} \cite{esrq2018}) to JSON \ref{json: search_request_body}.}} \label{fig: ms}
\end{figure}

\paragraph{Multimodal search} More often than not, scenarios more complicated than visual search will be encountered. For instance, a customer might be fascinated with the design and style of an armoire at her friend's house, but she might want to change its color to be better aligned with her own home design or want the price to be within her budget (see Figure \ref{fig: ms}). Searching using the picture snapped  is most likely in vain. To better enhance customers' shopping experiences, a visual search engine should be capable of retrieving results as {\em a joint outcome} by taking both the visual and textual requests from customers into consideration. Fortunately, our Elasticsearch-based visual system can immediately achieve this with one or two lines modifications in JSON \ref{json: search_request_body}. In particular, filters can be inserted within the  function score query to search only among products of customers' interests (e.g., within certain price range \cite{esrq2018}, attributes, colors). Moreover, general full-text query \cite{esft2018} can also be handled, score of which can be blended with the one from visual search in a weighted manner.  

\section{Vector to String Encoding}\label{sec: encoder}
The success of our approach hinges upon the quality of the encoder $\mc E(\cdot)$, which ideally should encourage closer vectors to share more sting tokens in common, so that the retrieval set $\mc R$ found based on token matching contains enough meaningful candidates. In the following, we first review the element-wise rounding encoder proposed by  \citet{rygl2017semantic}, and discuss its potential drawbacks. As a remedy, we propose a novel encoding scheme called subvector-wise clustering encoder.

%In the following, we will introduce two encoding schemes: {\em element-wise rounding} and {\em subvector-wise clustering}.

\subsection{Element-wise Rounding}
Proposed and examined by \citet{rygl2017semantic} and  \citet{ruuvzivckaflexible}, the element-wise rounding encoder rounds each value in the numerical vector to $p$ decimal places (where $p\ge 0$ is a fixed integer), and then concatenates its positional information and rounded value as the string tokens. 

\begin{example}\label{ex: ewr}
For a vector $\bm x = [0.1234, -0.2394, 0.0657]$, rounding to two decimal places (i.e.,  $p=2$) produces string tokens of $\bm x$ as 
\[
\bm s =  \set{\mbox{``pos1val0.12''},\mbox{``pos2val-0.24''}, \mbox{``pos3val0.07''}}.
\]
The encoded positional information is essential for the inverted-index-based search system to match (rounded) values at the same position without confusion. Suppose on the other hand, positional information is ignored, and thus
\[
\bm s =  \set{\mbox{``val0.12''},\mbox{``val-0.24''}, \mbox{``val0.07''}}.
\]
Then the attribute ``val0.12'' could be mistakenly matched by another encoded token even when it is not produced from the first entry.
\end{example}
%\cm{why positional info.? to match values for the same position...}

For a high-dimensional vector $\bm x \in \reals^d$, this vanilla version of the element-wise rounding encoder will generate a large collection of string tokens (essentially with $|\mc E(\bm x)| = d$), which makes it infeasible for Elasticsearch to compute \eqref{eqn: retrieval} in real time. 

\paragraph{Filtering} As a remedy, \citet{rygl2017semantic} presents a useful filtering technique to sparsify the string tokens. In specific, only top-$m$ entries in terms of magnitude are selected to create rounding tokens. 

\begin{example}
For the same setting with Example \ref{ex: ewr}, when $m$ is set as $2$, the string tokens will be produced as 
\[
\bm s =  \set{\mbox{``pos1val0.12''},\mbox{``pos2val-0.24''}}
\]
with only the first and second entries being selected; 
and when $m$ is set as $1$, the string tokens will be produced as 
\[
\bm s =  \set{\mbox{``pos2val-0.24''}},
\]
with only the second entry being selected.
\end{example}
 
\paragraph{Drawbacks}
Although the filtering strategy is suggested to maintain a good balance between feature sparsity and search quality \cite{rygl2017semantic, ruuvzivckaflexible}, it might not be the best practice to reduce the number of string tokens with respect to finding nearest neighbors in general. First, for two points $\hat{\bm x} , \bm x \in \reals^d$, their Euclidean distance
\begin{flalign}\label{eqn: ed}
\norm{\hat{\bm x} - \bm x}{2}^2 = \sum_{i=1}^d (\hat x_i - x_i)^2,
\end{flalign}
is summed along each axis equally rather than biasedly based on the magnitude of $\hat x_i$ (or $x_i$). In specific, a mismatch/match with a (rounded) value $0.01$ does not imply that it is less important than a mismatch/match with a $0.99$, in terms of their contributions to the sum \eqref{eqn: ed}.  What essentially matters is the deviation $\Delta_i := \hat x_i - x_i$ rather than the value of $\hat x_i$ (or $x_i$) by itself. Therefore, entries with small magnitude should not be considered as less essential and be totally ignored.  Second,  the efficacy of the filtering strategy is vulnerable to data distributions. For example, when the embedding vectors are binary codes \cite{heinly2012comparative, lai2015simultaneous, lu2017deep, song2017binary, loncaric2018convolutional}, choosing top-$m$ entries will lead to an immediate tanglement (see \cite{mu2019empowering} for a recently proposed implementation on Elasticsearch to efficiently conduct exact nearest neighbor search in Hamming space).

In the next subsection, we will propose an alternative encoder, which keeps all value information into consideration and is also more robust with respect to the underlying data distribution.

\subsection{Subvector-wise Clustering}
\begin{figure}[ht]
	\centerline{\includegraphics[width=3in]{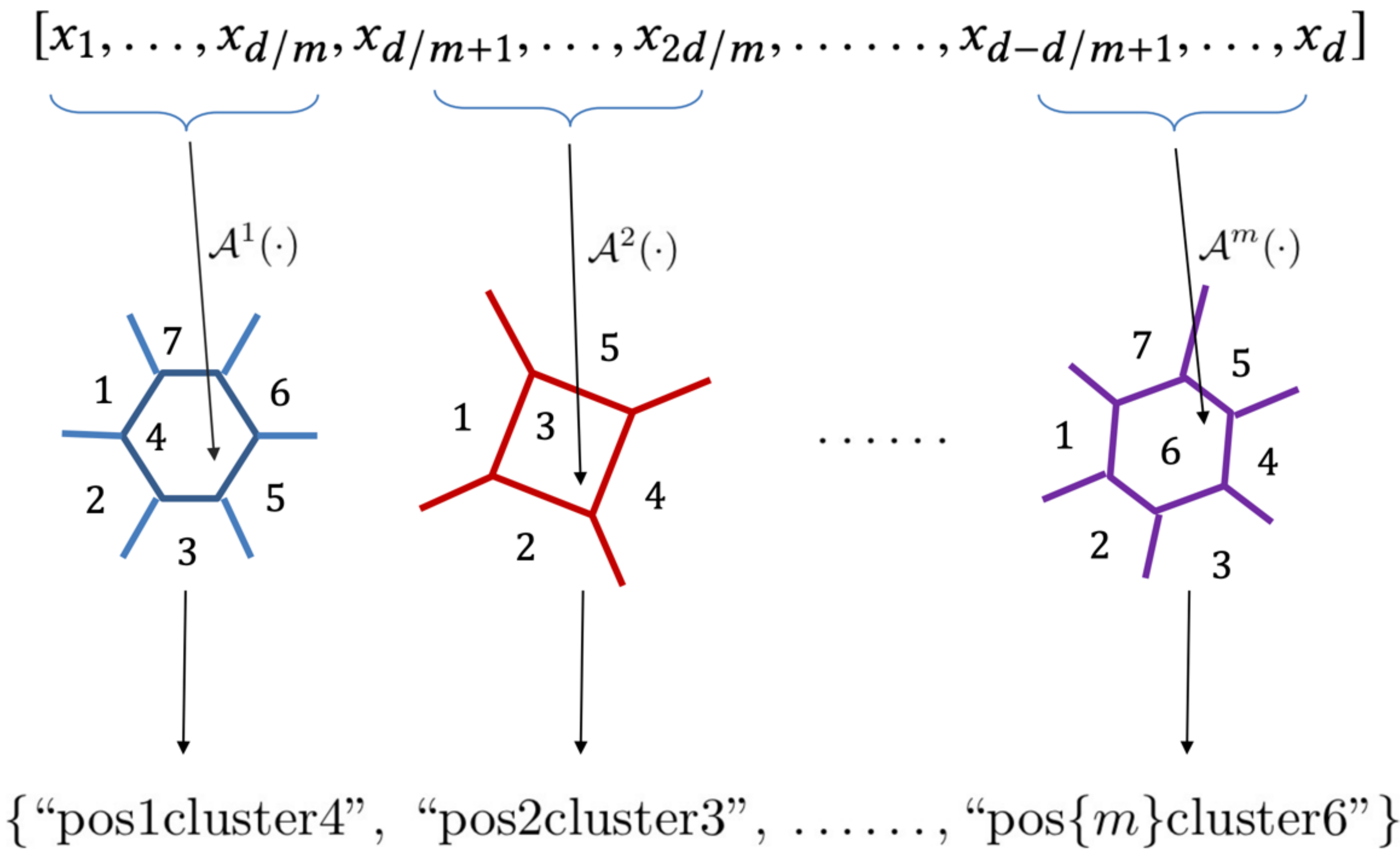}}
	\caption{{\bf Illustration of the subvector-wise clustering encoder.} \normalfont{The vector $\bm x \in \reals^d$ is divided into $m$ subvectors. Subvectors at the same position are considered together to be classified into $k$ clusters. Then each subvector is encoded into a string token by combining its position in $\bm x$ and the cluster it belongs to, so exact $m$ string tokens will be produced.}} \label{fig: swc_encoder}
\end{figure}

Different from the element-wise rounding one, an encoder that operates on a subvector level will be presented in this part. The idea is also quite natural and straightforward. For any vector $\bm x \in \reals^d$, we divide it into $m$ subvectors\footnote{For simplicity, we assume $m$ divides $d$.}, 
\begin{flalign}\label{eqn: divide_vector}
[\underbrace{x_1, \ldots, x_{d/m}}_{\bm x^1}, \underbrace{x_{d/m+1}, \ldots, x_{2d/m}}_{\bm x^2}, \ldots \ldots, \underbrace{x_{d-d/m+1}, \ldots, x_{d}}_{\bm x^m}].
\end{flalign}
Denote $\mc X^i:=\set{\bm x_1^i, \bm x_2^i, \ldots, \bm x_n^i}$ as the collection of the $i$-th subvectors from $\mc X$ for $i = 1,2,\ldots, m$. We will then separately apply the classical $k$-means algorithm \cite{macqueen1967some} to divide each $\mc X^i$ into $k$ clusters with the learned assignment function 
\[
\mc A^i: \reals^{d/m} \to \set{1,2, \ldots, k}
\]
assigning each subvector to the cluster index it belongs to. Then for any $\bm x \in \reals^d$, we will encode it into a collection of $m$ string tokens
\begin{flalign}
\set{\mbox{``pos1cluster\{$\mc A^1(\bm x^1)$\}''}, 
\mbox{``pos2cluster\{$\mc A^2(\bm x^2)$\}''}, \ldots}.
\end{flalign}
The whole idea is illustrated in Figure \ref{fig: swc_encoder}. The trade-off between search latency and quality is well controlled by the parameter $m$. In specific, a larger $m$ will tend to increase the search quality as well as the search latency, as more string tokens per each vector will be indexed.

In contrast with the element-wise rounding encoder, our subvector-wise clustering encoder obtains $m$ string tokens without throwing away any entry in $\bm x$, and will generate string tokens more adaptive with the data distribution, as the assignment function $\mc A^i(\cdot)$ for each subspace is learned through $\mc X^i$ (or data points sampled from $\mc X^i$).

\section{Experiment}\label{sec: experiment}
In this section, we will compare the performance of the subvector-wise clustering encoder and the element-wise rounding one in terms of both precision and latency, when they are being used in our content-based image retrieval system built upon Elasticsearch.

\paragraph{Settings} Our image datasets consists of around half a million images selected from Jet.com's furniture catalog \cite{jetfurniture2018}. For each image, we extract its image feature vector using the pretrained \textsc{Inception-ResNet-V2} model \cite{szegedy2017inception}. In specific,  each image is embedded into a vector in $\reals^{1536}$ by taking the output from the penultimate layer (i.e., the last average pooling layer) of the neural network model. String tokens are produced respectively with encoding schemes at different configurations. For the element-wise rounding encoder,  we select $p \in \set{0,1,2,3}$, and $m \in \set{32, 64, 128, 256}$. For the subvector-wise clustering encoder, we experiment with  $k \in \set{32,64, 128, 256}$ and $m \in \set{32, 64, 128, 256}$. Under each scenario, we  index the image feature vectors and their string tokens into a single-node Elustersearch cluster deployed on a Microsoft Azure virtual machine \cite{msaz2018} with 12 cores and 112 GiB of RAM. To better focus on the comparison of the efficacy in 
encoding scheme, only vanilla setting of Elasticsearch (one shard and zero replica) is used in creating each index. 

\paragraph{Evaluation} 
To evaluate the two encoding schemes, we randomly select 1,000 images to act as our visual queries. For each of the query image, we find the set of its 24 nearest neighbors in Euclidean distance, which is treated as gold standard. We use Precision@24 \cite{schutze2008introduction}, which measures the overlap between the 24 images retrieved from Elasticsearch (with $r \in \set{24, 48, 96, \ldots, 6144}$ respectively) and the gold standard, to evaluate the retrieval efficacy of different encoding methods under various settings. We also record the latency for Elasticsearch to execute the retrieval and reranking steps in the searching phase. 

\paragraph{Results}

\begin{figure}
	\centering
	\includegraphics[width=8cm]{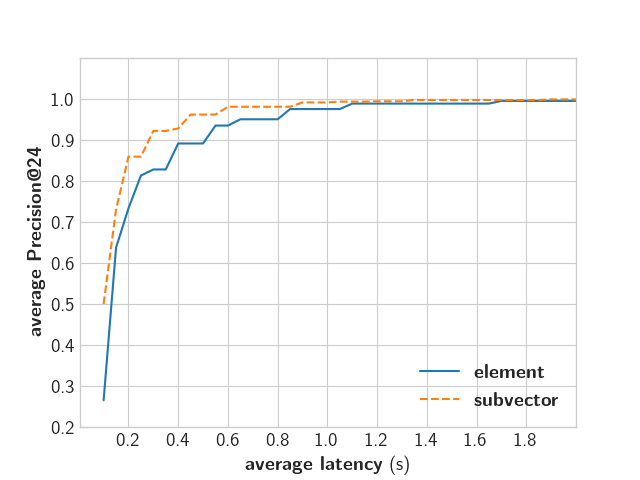}
	\caption{{\bf Pareto frontier for the element-wise rounding and the subvector-wise clustering encoders in the space of latency and precision.} \normalfont{It can be clearly seen that our subvector-wise encoding scheme is capable of achieving higher precision with smaller latency.}}
	\label{fig: major_result}
\end{figure}

In Table \ref{Tab: precision}, we report the Precision@24 and search latency averaged over the 1,000 queries randomly selected. Results corresponding to $p \in \set{2, 3}$ or $r \in \set{24, 48}$ are skipped as they are largely outperformed by other settings.  Configurations that can achieve precision $\ge 80\%$ and latency $\le 0.5$s are highlighted in bold. From Table  \ref{Tab: precision}, we can see that the subvector-wise encoder outperforms the element-wise one, as for  all results obtained by the element-wise encoder, we can find a better result from the subvector-wise one in both precision and latency. To better visualize this fact, we plot the {\em Pareto frontier curve} over the space of precision and latency in Figure \ref{fig: major_result}. In specific, the dashed (resp. solid) curve in Figure \ref{fig: major_result} is plotted as the best average Precision@24 achieved among all configurations we experiment for element-wise rounding (resp. subvector-wise clustering) encoder, under different latency constraints.  From Figure \ref{fig: major_result}, we can more clearly observe that the subvector-wise encoder surpasses the element-wise one. Notably, when we require the search latency to be smaller than 0.3 second, the subvector-wise encoder is able to achieve an average Precision@24 as 92.14\%, yielding an improvement of more than 11\% over the best average Precision@24 that can be obtained by the element-wise one. 

\section{Future Work}
Although our subvector-wise clustering encoder outperforms the element-wise rounding one, it might be still restrictive to enforce a vector to be divided into subvectors exclusively using \eqref{eqn: divide_vector}, which could potentially downgrade the performance of the encoder. Our next step is to preprocess the data (e.g., transform the data through some linear operation $\bm x \mapsto  \mc T[\bm x]$ with $\mc T[\cdot]$ learned from the data) before applying our subvector-wise clustering encoder. We believe this flexibility will make our encoding scheme more robust and adaptive with respect to different image feature vectors extracted from various image descriptors. Another interesting research direction is to evaluate the performances of different encoding schemes in other information retrieval contexts--e.g., {\em neural ranking model based textual searches} \cite{shen2014latent, mitra2017neural, Brenner2018ete}, where relevances between user-issued queries and catalog products are modeled by their Euclidean distances in the embedding space to better match customers' intents with products.

\begin{center}
	\begin{table*}[h]
		{ 
			\hfill{}
			\begin{tabular}{|C{.75cm}| C{1.5cm} |C{2.5cm}|C{2.1cm}|C{2.1cm}|C{2.1cm}|C{2.1cm}|}
				\hline
				\multirow{2}{.75cm}{\centering $r$} &
				\multirow{2}{1.5cm}{\centering Encoding} &
				\multirow{2}{2.5cm}{\centering Round./Cluster.}& \multicolumn{4}{p{8.4 cm}|}{\centering \# of feature tokens ($m$)} \\
				\cline{4-7} & & & \multicolumn{1}{c|}{32}  & \multicolumn{1}{c|}{64} & \multicolumn{1}{c|}{128} & \multicolumn{1}{c|}{256}\\ 
				\hline
				%				\hline
				%				48
				%				& element    &0-decimal-place & 42.87\%  |  0.1283  &  52.74\% |  0.2382 & 64.42\% | 0.5298  & 78.56\% | 1.5387    \\
				%				& element    &1-decimal-place & 21.26\%  |  0.0955   &  30.44\% | 0.1629  & 41.34\% | 0.3372  & 53.61\% | 0.8146  \\
				%				\cdashline{2-7}
				%				& subvector &32-centroids & 27.21\% | 0.1132 & 43.62\% | 0.1962  & 62.75\% | 0.3825 & 76.88\% | 0.8982  \\
				%				& subvector &64-centroids & 30.63\% | 0.1005 & 47.16\% | 0.1683 & 64.72\% | 0.3481  & 76.65\% | 0.7646  \\
				%				& subvector &128-centroids & 34.91\% | 0.0960 & 50.28\% | 0.1595 & 67.19\% | 0.3135  & 75.25\% | 0.7250  \\
				%				& subvector &256-centroids & {38.71\% | 0.0954} & 53.97\% | 0.1531 & 67.43\% | 0.3016 & 72.57\% | 0.6845  \\
				%				\hline
				\hline
				96
				& element    &0-decimal-place & 53.43\%  |  0.1237  &  64.35\% |  0.2339 & 76.44\% | 0.5256  & 88.64\% | 1.5342    \\
				& element    &1-decimal-place & {26.56\%  |  0.0920}   &  37.94\% | 0.1592  & 50.35\% | 0.3370  & 63.71\% | 0.8207  \\
				\cdashline{2-7}
				& subvector &32-centroids & 34.80\% | 0.1111 & 54.45\% | 0.1914  & 74.22\% | 0.3760 & 87.44\% | 0.8914  \\
				& subvector &64-centroids & 39.52\% | 0.0963 & 58.51\% | 0.1630 & 76.70\% | 0.3426  & 87.28\% | 0.7563  \\
				& subvector &128-centroids & 44.43\% | 0.0914 & 61.93\% | 0.1544 & 78.89\% | 0.3088  & 85.58\% | 0.7186  \\
				& subvector &256-centroids & 50.00\% | 0.0900 & 66.22\% | 0.1480 & 79.05\% | 0.2970  & 82.89\% | 0.6757  \\
				\hline
				\hline
				192
				& element    &0-decimal-place & {63.72\%  |  0.1405}  &  74.63\% |  0.2499 & 85.38\% | 0.5416  & 94.13\% | 1.5536    \\
				& element    &1-decimal-place & 32.49\%  |  0.1084  &  45.50\% | 0.1748  & 59.05\% | 0.3529  & 72.12\%  | 0.8424  \\
				\cdashline{2-7}
				& subvector &32-centroids & 43.73\% | 0.1256 & 64.88\% |  0.2080  & \textbf{83.13\%} | \textbf{0.3917} & 93.56\% | 0.9146  \\
				& subvector &64-centroids & 48.84\% | 0.1130 & 69.14\% | 0.1795 & \textbf{85.12\% | 0.3594}  & 93.28\% | 0.7745  \\
				& subvector &128-centroids & 55.14\% | 0.1082 & 72.62\% | 0.1714 & \textbf{87.08\% | 0.3250}  & 91.97\% | 0.7367 \\
				& subvector &256-centroids & {61.41\% | 0.1066} & {77.08\% | 0.1644} & \textbf{87.32\% | 0.3137}  & 89.28\% | 0.6915  \\
				\hline
				\hline
				384
				& element    &0-decimal-place & {73.30\%  |  0.1749}  &  \textbf{82.76\% |  0.2852} & {91.19\% | 0.5756}  & 97.03\% | 1.5963    \\
				& element    &1-decimal-place & 38.94\%  |  0.1431   &  53.43\% | 0.2093  & 67.12\% | 0.3877  & 79.25\%  | 0.8741  \\
				\cdashline{2-7}
				& subvector &32-centroids & 53.37\% | 0.1603 & 73.92\% | 0.2417  & \textbf{89.06\%} | \textbf{0.4262} & 96.82\% | 0.9509   \\
				& subvector &64-centroids & 59.01\% | 0.1479 & 78.15\% | 0.2139 & \textbf{91.25\%} | \textbf{0.3935}  & 96.59\% | 0.8097  \\
				& subvector &128-centroids & 66.20\% | 0.1433 & \textbf{81.56\% | 0.2061} & \textbf{92.75\% | 0.3596}  & 95.44\% | 0.7705  \\
				& subvector &256-centroids & 73.01\% | 0.1415 & \textbf{85.88\% | 0.1995} & \textbf{92.67\% | 0.3520}  & 93.38\% | 0.7243  \\
				\hline
				\hline
				768
				& element    &0-decimal-place & \textbf{81.27\%  |  0.2455}  &  \textbf{89.09\% |  0.3547} &  {94.98\% | 0.6443}  & {98.60\% | 1.6613 }   \\
				& element    &1-decimal-place & 45.83\%  |  0.2130   &  61.30\% | 0.2801  & 74.60\% | 0.4574  & 84.87\%  | 0.9427  \\
				\cdashline{2-7}
				& subvector &32-centroids & 63.45\% | 0.2297 & \textbf{81.30\% | 0.3117}  & \textbf{93.40\%} | \textbf{0.4974} & 98.58\% | 1.0195  \\
				& subvector &64-centroids & 69.01\% | 0.2182 & \textbf{85.47\% | 0.2837} & \textbf{95.41\%} | \textbf{0.4647}  & {98.38\% | 0.8798}  \\
				& subvector &128-centroids & 76.70\% | 0.2133 & \textbf{88.91\% | 0.2762} & \textbf{96.13\%} | \textbf{0.4288}  & 97.50\% | 0.8402  \\
				& subvector &256-centroids & \textbf{83.55\% | 0.2112} & \textbf{92.14\% | 0.2701} & \textbf{95.90\%} | \textbf{0.4267}  & 95.94\% | 0.7970  \\
				\hline
				\hline
				1536 %done
				& element    &0-decimal-place & \textbf{87.55\%  |  0.3923}  &   93.45\% |  0.5027 &  {97.47\% |  0.8012}  & 99.29\% | 1.8486   \\
				& element    &1-decimal-place & 53.76\%  |  0.3656   &  68.68\% | 0.4361  & 81.05\% | 0.6069  & 89.48\%  | 1.0931  \\
				\cdashline{2-7}
				& subvector &32-centroids & 72.75\% | 0.3703 & \textbf{87.30\%} | \textbf{0.4524}  & 96.14\% | 0.6400 & {99.36\% | 1.1574}  \\
				& subvector &64-centroids & 78.85\% | 0.3581 & \textbf{91.52\%} | \textbf{0.4218} & 97.74\% | 0.6045  & {99.28\% | 1.0188}  \\
				& subvector &128-centroids & \textbf{86.00\% | 0.3537} & \textbf{94.12\% | 0.4158} & {98.03\% | 0.5665} & {98.60\% | 0.9763}  \\
				& subvector &256-centroids & \textbf{91.16\%} | \textbf{0.3512} & \textbf{95.97\% | 0.4087} & 97.70\% | 0.5582  & 97.44\% | 0.9281  \\
				\hline
				\hline
				3072
				& element    &0-decimal-place &  92.38\% | 0.6843   &   96.40\% | 0.8166   &  98.80\% | 1.0909 & 99.63\% | 2.1638    \\
				& element    &1-decimal-place &  61.50\% | 0.6625  & 75.62\% | 0.7380  &  86.32\% | 0.9135  & 92.85\% | 1.3946    \\
				\cdashline{2-7}
				& subvector &32-centroids & 81.25\% | 0.6645   &  92.11\% | 0.7483   & 97.95\%  | 0.9375  & 99.68\% | 1.4589    \\
				& subvector &64-centroids &  87.82\% | 0.6556   &  96.32\%  | 0.7131   &  99.00\% | 0.9006    & 99.68\% | 1.3189    \\
				& subvector &128-centroids &  93.26\% | 0.6508  &  97.72\%  | 0.7126   &  99.08\% | 0.8604  & 99.21\% | 1.2756    \\
				& subvector &256-centroids & 96.06\% | 0.6470   &  97.94\%  | 0.7074   &  98.72\% | 0.8566  & 98.37\% | 1.2230    \\
				\hline
				\hline
				6144
				& element    &0-decimal-place & 95.52\% | 1.2630  &  98.22\%  | 1.3778   &  99.45\% | 1.6737  & 99.82\% | 2.7669    \\
				& element    &1-decimal-place &  68.26\% | 1.2535 &   81.75\% | 1.2942  &  90.69\% | 1.4800    &  95.24\% | 1.9542    \\
				\cdashline{2-7}
				& subvector &32-centroids & 89.61\% | 1.2081  &  95.86\%  | 1.2938   &  99.10\% | 1.4892  & 99.85\% | 2.0124    \\
				& subvector &64-centroids & 95.43\% | 1.2031   &  98.87\%  | 1.2537   &  99.65\% | 1.4459 & 99.82\% | 1.8647    \\
				& subvector &128-centroids &  97.56\% | 1.1985   &   99.13\% | 1.2565   &  99.54\% | 1.3959 & 99.52\% | 1.8200    \\
				& subvector &256-centroids &  98.20\% | 1.1957   &   98.90\% | 1.2542   &  99.25\% | 1.4037  & 98.97\% | 1.7586    \\
				\hline			
			\end{tabular}
		}
		\hfill{}
		\caption{Mean Precision@24 | ES average latency. {\normalfont For each setting, we average the Precision@24 and the number of seconds used  over the 1,000 query images randomly selected from the furniture dataset. Settings with mean precision $\ge 80\%$ and latency $\le 0.5s$ are highlighted in bold.
		}}
		\label{Tab: precision}
	\end{table*}
\end{center}

\section*{Acknowledgement}
We are grateful to three anonymous reviewers for their helpful suggestions and comments that substantially
improve the paper. We would also like to thank Eliot P. Brenner and Aliasgar Kutiyanawala for proofreading the first draft of the paper.

\bibliographystyle{ACM-Reference-Format}
\bibliography{vss}

\end{document}